\documentclass{article}

\PassOptionsToPackage{numbers, compress}{natbib}
\usepackage{amsthm}
\usepackage{amssymb}
\usepackage[preprint]{neurips_2023}

\usepackage[utf8]{inputenc} % allow utf-8 input
\usepackage[T1]{fontenc}    % use 8-bit T1 fonts
\usepackage{hyperref}       % hyperlinks
\usepackage{url}            % simple URL typesetting
\usepackage{booktabs}       % professional-quality tables
\usepackage{amsfonts}       % blackboard math symbols
\usepackage{nicefrac}       % compact symbols for 1/2, etc.
\usepackage{microtype}      % microtypography
\usepackage{xcolor}         % colors
\usepackage{graphicx}
\usepackage{amsmath}
\usepackage{float}

\title{RLInspect: An Interactive Visual Approach to Assess Reinforcement Learning Algorithm}

\author{
  Geetansh Kalra \\
  Engineering for Research \\
  Thoughtworks Technologies \\
  Pune, MH, India \\
  \texttt{geetansh.kalra@thoughtworks.com}
  \And
  Divye Singh \\
  Engineering for Research \\
  Thoughtworks Technologies \\
  Pune, MH, India \\
  \texttt{divye.singh@thoughtworks.com}
  \And
  Justin Jose \\
  Engineering for Research \\
  Thoughtworks Technologies \\
  Pune, MH, India \\
  \texttt{justinj@thoughtworks.com,}
}

\begin{document}

\maketitle

\begin{abstract}
Reinforcement Learning (RL) is a rapidly growing area of machine learning that finds its application in a broad range of domains, from finance and healthcare to robotics and gaming.  Compared to other machine learning techniques, RL agents learn from their own experiences using trial and error, and improve their performance over time. However, assessing RL models can be challenging, which makes it difficult to interpret their behaviour. While reward is a widely used metric to evaluate RL models, it may not always provide an accurate measure of training performance. In some cases, the reward may seem increasing while the model's performance is actually decreasing, leading to misleading conclusions about the effectiveness of the training. To overcome this limitation, we have developed RLInspect - an interactive visual analytic tool, that takes into account different components of the RL model - state, action, agent architecture and reward, and provides a more comprehensive view of the RL training. By using RLInspect, users can gain insights into the model's behaviour, identify issues during training, and potentially correct them effectively, leading to a more robust and reliable RL system.
\end{abstract}
\section{Introduction}
\label{sec::into}

Machine learning systems have made impressive advances due to their ability to learn from high dimensional, non-linear, and imbalanced data \cite{LeCun2015-yj}. However, learning from complex data often leads to complex models, which require rigorous evaluation. Over the time, lot of performance metrics \cite{blagec2021critical, Zhou2021-fd} and visualisation tools \cite{dataflow_graph_tensorflow,vega_lite,kahng2017cti,interacting_with_predictions,yosinski2015understanding} have been proposed and are being used in different supervised learning settings. These metrics and tools have helped in qualifying performance and understanding inner working of supervised learning models. 

Reinforcement learning (RL) is a branch of ML where the agent (decision making entity) interacts with the environment and learns based on the reward (feedback) received. In recent years RL has seen an increase in real-world interest and applications due to its ability to interact with the environment, making it a valuable tool for addressing real-world problems across multiple domains \cite{rl_popularity}. It is increasingly being used to solve tasks from different areas like everyday life assistance \cite{li2019reinforcement,chatgpt_rlhf,liu2023summary}, policy making \cite{ai_economist,tilbury2022reinforcement,kwak_ling_hui_2021,khadilkar2020optimising} and high-stakes domain like finance \cite{hambly2021recent} and clinical decision systems \cite{rl_clinical_decision_system}. However, there is a lack of consistent metrics making it difficult to assess true performance of RL \cite{NEURIPS2020_48db7158}. Due to the complexity of the real-world problems, selecting a suitable metric for RL becomes even more important, as flawed or incomplete metric can lead to inconsistent performance and reproducibility \cite{jordan2020evaluating}.

A suitable and robust metric can either be a numerical summary or can be in the form of a visual representation. Although numerical metrics can summarise the performance in a few numbers, visual tools can often provide a more intuitive and immediate understanding of complex information \cite{tufte_2018}. This is because visual representations can convey patterns, trends, and relationships in data and can also help keep track of what manipulations are being applied at what stage. While static images can offer analytical insights, their effectiveness is constrained when dealing with complex data or representations \cite{jisoo2007interaction}. Therefore, interactiveness plays an essential role in a visual tool for providing the flexibility to focus on the different regions of the visualisation, as and when needed.

An RL training process consists of three main components which are input states, actions and reward and also neural network in case of deep RL. In this paper, we present RLInspect, an interactive visual tool for understanding and potentially debugging the RL training process. It provides different methods to assess the behaviour of RL algorithms with respect to the aforementioned components during the training process, which can help evaluate the quality of the RL training process and in-turn, the trained model.
\section{Related Work}
\label{sec::rel-work}

The recent interest in understanding, explaining and interpreting the RL models has led to development of analysis methods and techniques. \cite{Wesel2017ChallengesIT} in their report talk about the need for verifying RL and what can be verified offline and during runtime. This section categorises previous work into two groups - one which focuses on explaining RL and the other, where a visual tool is developed for understanding RL algorithms.

\textbf{Explaining RL}: There has been some work done on explaining the behaviour of the RL model. \cite{iyer2018obj_saliency_map} proposes an object saliency map to understand the extent of influence various objects in an image have on the q-values. Their method is limited to images as input and depends on the user’s understanding of the best action given the object saliency map, which is not always possible. There is also some work done on measuring reliability of RL algorithms \cite{chan2020measuring}, which looks at conditional value-at-risk and dispersion of the cumulative rewards, both during training and testing. This can then be used to analyse results and compare different algorithms. Another work proposes carrying out multiple trials followed by calculating two metrics. First is performance percentile using empirical cumulative distribution functions of average returns and second is calculating confidence intervals using their proposed method called performance bound propagation \cite{jordan2020evaluating}. In another work, \cite{Sequeira_2020} extracted the interestingness element for explaining RL by introducing the introspection framework. This framework looks at the frequency state and state-action pair, execution certainty, transition value and state-action sequence from local minima to local maxima. This idea was further extended in XRL-DINE which enhanced and combined interestingness elements and rewards decomposition to explain self-adaptive systems \cite{feit2022explaining}.

\textbf{Visual tools}: The increasing interest in visual analysis methods have led to development of many visual tools for analysing RL algorithms. MDPVis provides simulator-visualization interface, generalized MDP information visualization and exposes model components to the user \cite{MCGREGOR201793}. A visual analytic tool called DQNViz is presented in \cite{wang2019dqnviz} which provides visual analytics in the form of four coordinated views - statistics view, epoch view, trajectory view and segment view. Later, \cite{jaunet2020drlviz} presented DRLViz which is a visual analytics interface that focuses on understanding connection between latent memory representation and decisions of a trained agent. The authors have used VizDoom to demonstrate capabilities of DRLViz in identifying elements responsible for low-level decision making and high-level strategies.
\section{RLInspect}
\label{sec::rlinspect}

RLInspect is an intuitive and user-friendly tool that aims to improve the performance of Reinforcement Learning models by providing users with a comprehensive understanding of their model's behaviour. It offers a range of interactive visual plots to analyze and gain insights into potential issues. Additionally, RLInspect can work with both continuous and discrete action and state-spaces, making it a versatile tool for a wide range of RL applications.

\subsection{Architecture of RLInspect}
\begin{figure}[H]
    \centering
    \includegraphics[width=0.5\textwidth]{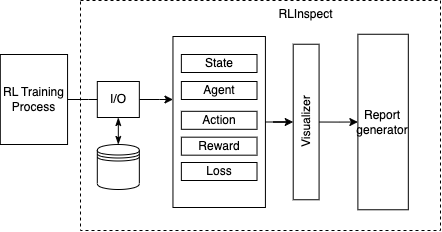}
    \caption{Architecture of RLInspect}
    \label{fig::rlinspect_architecture}
\end{figure}

\begin{figure}[H]
    \begin{center}
        \includegraphics[width=0.9\textwidth, height=0.5\textwidth]{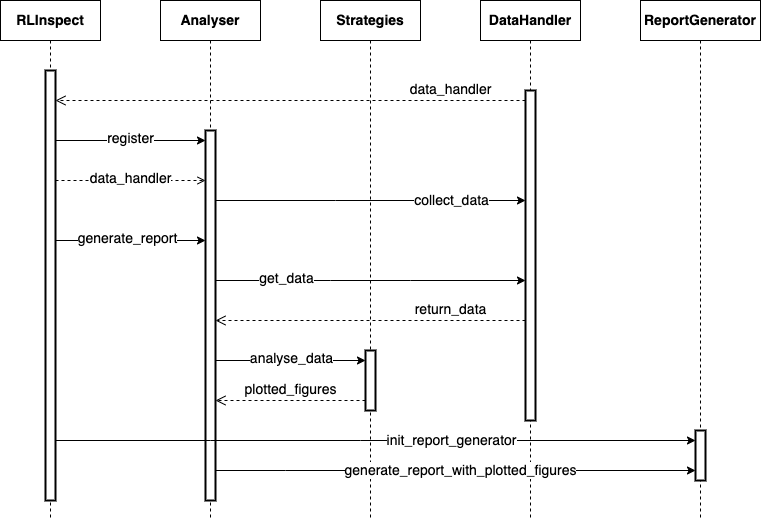}
        \caption{Sequence of data flow in RLInspect}
        \label{fig::rlinspect_sequence_diagram}
    \end{center}
\end{figure}

RLInspect is designed with a plug-able architecture, empowering users to select the most suitable modules for their training purposes or even customize the modules and their analysis to cater to their unique requirements. It is comprised of three main components: I/O, Analyzers, and Report Generator (Figure \ref{fig::rlinspect_architecture}). The I/O is responsible for managing all input and output operations and is equipped with a central hub for data management known as the Data Handler. Upon initialization of RLInspect, this \texttt{DataHandler} is created to manage data flow. The \texttt{Analyzers}, on the other hand, are created and registered in the Analyzers Registry of RLInspect, though they can also be used independently. Through the use of the \texttt{DataHandler}, the \texttt{Analyzers} gather the necessary data. Each analyzer is equipped with its own analysis to extract insights from the collected data. To enhance interpretability, each analyzer's analysis generates interactive plots using Plotly \cite{plotly}, providing a more user-friendly and comprehensive visualization of the results. Finally, the \texttt{ReportGenerator} aggregates all the generated plots into a single HTML file for visualization. This process flow is shown in Figure \ref{fig::rlinspect_sequence_diagram}.

% Furthermore, RLInspect has a plug-able architecture that empowers users to select the most suitable modules for their training purposes, or even customize the modules and their analysis to cater to their unique requirements.

In the upcoming sections we will present an overview of each module of RLInspect and demonstrate how they provided insights for the RL agent trained on a Cartpole environment of OpenAI Gym \cite{brockman2016openai}.

% In the upcoming section, we will showcase how RLInspect's modules provide insights for an agent trained on the CartPole environment of OpenAI Gym \cite{brockman2016openai}. We will present an overview of each module and demonstrate how they contribute to a comprehensive analysis of the agent's performance.

% In the upcoming sections, we will provide an analysis of the effectiveness of DQN in the CartPole environment of OpenAI Gym \cite{brockman2016openai} using RLInspect. We will present an overview of each module of RLInspect and demonstrate how they provided insights for this particular use case.

\subsection{State Module}
\label{subsec:state}
\begin{figure}[htb]
    \includegraphics[width=0.9\textwidth]{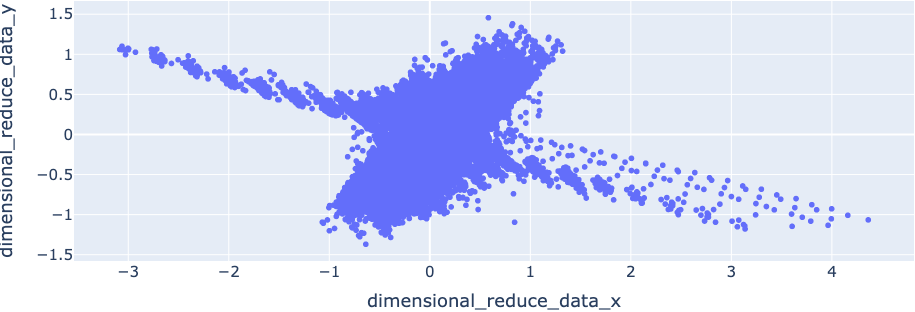}
    \caption{State-space distribution in Cartpole.}
    \label{fig: state-space Distribution in Cartpole}
\end{figure}

The State Module in RLInspect is designed to assist users in gaining a comprehensive understanding of the state-space on which their RL model is being trained. This module utilizes scatter plots to provide users with insights into the distribution of the state-space, which may include a combination of exploration-exploitation trade-off states, as well as specific states that were sampled during training and their corresponding locations within the state-space. However, visualizing state-spaces can be particularly challenging when state variables are multidimensional and can take on values that are either continuous or discrete. To overcome this challenge, RLInspect utilizes Incremental Principal Component Analysis (IPCA) \cite{Ross2007} which allows the visualization of high-dimensional states in a lower-dimensional space. State Module offers three data distribution analysis: 

\paragraph{State-space Distribution:}
This analysis is used for visualizing the distribution of states which provides valuable insights into the underlying structure of the state-space and assess the quality of the data coverage. Figure \ref{fig: state-space Distribution in Cartpole} shows the state-space explored by the RL agent for the cartpole environment.

\begin{figure}[H]
    \includegraphics[width=0.9\textwidth]{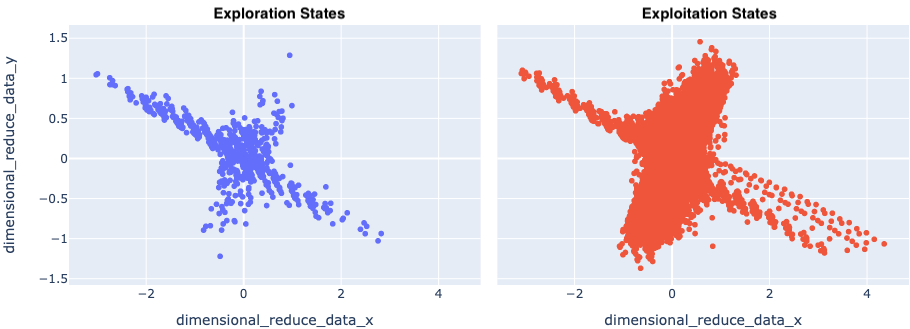}
    \caption{Exploration vs Exploitation in Cartpole.}
    \label{fig: Exploration vs Exploitation in Cartpole}
\end{figure}

\paragraph{Exploration vs Exploitation Distribution:}
Exploration-exploitation trade-offs are critical in RL, as they dictate the agent's behaviour in an uncertain and dynamic environment. Agents must balance the exploration of new state-action pairs with exploiting the current knowledge to achieve their goals. Identifying areas in the state-space where the model over or under explores, can help fine-tune algorithms and improve overall performance. To check this trade-off, RLInspect generates two facet columns within each scatter plot, allowing for side-by-side comparisons of the exploration and exploitation states. In Figure \ref{fig: Exploration vs Exploitation in Cartpole}, it can be observed that the cartpole-agent has under-explored the state-space. 

\begin{figure}[H]
    \includegraphics[width=0.9\textwidth]{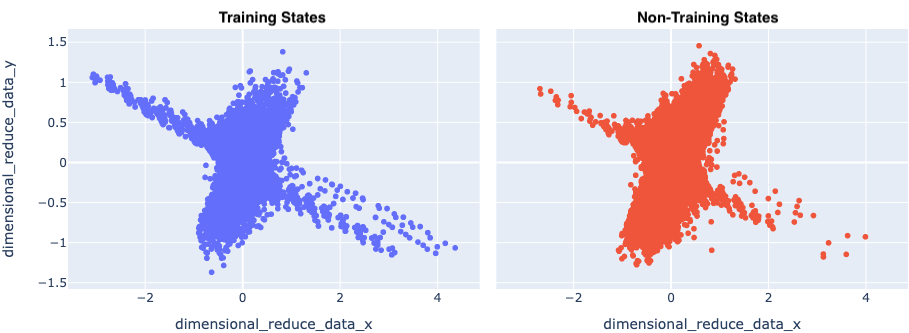}
    \caption{Training vs Non-Training states in Cartpole.}
    \label{fig: Training vs Non-Training in Cartpole}
\end{figure}

\paragraph{Training States Distribution:}
Analyzing the training state distribution helps identify if the model is being uniformly trained across the state-space or if there is a bias towards a region. This is important because if the model is biased towards a particular region, it may not perform well when making predictions for other regions. As observed in figure \ref{fig: Training vs Non-Training in Cartpole}, the distribution of training states matches closely with the non-training states. This suggests that the model has been effectively trained on all the states it has encountered, enabling it to make accurate predictions.

\subsection{Action Module}
\label{subsec:action}
The action module focuses on the behaviour of RL algorithm's decision making. Looking at how the decision making evolves through the training run gives insight into stability and confidence of the actions. All the analyses in the action module are carried out on a predefined set of $\textit{k}$ states and the average across their analysis is taken as a single number representative of the model behaviour. The action module consists of following analysis techniques:

\paragraph{Action confidence:}
This analysis looks at the confidence of an action taken by the RL algorithm. The confidence of an action is defined as the function of policy distribution, $\pi(s)$, which is the probability density function or probability mass function of action for a given state.

\begin{equation}
\label{eq::confidence}
    \text{confidence}(s) = 1 - \text{entropy}(\pi(s))
\end{equation}
For well behaved RL algorithm, the confidence of an action converges to one as the training converges.

\begin{proof}
The update rule for action value in TD(0)is given as: 
\begin{equation}
\label{eq::q_update}
    Q_{t+1}^\pi(s_t, a_t) = Q_{t}^\pi(s_t, a_t) + \alpha(\mathcal{R}_t + \gamma \max_{a} Q_{t}^\pi(s_{t+1}, a) - Q_{t}^\pi(s_t, a_t))
\end{equation}

where, $\mathcal{R}_t + \gamma \max_{a} Q_{t}^\pi(s_{t+1}, a)$ is the target and $\mathcal{R}_t + \gamma \max_{a} Q_{t}^\pi(s_{t+1}, a) - Q_{t}^\pi(s_t, a_t)$ is the error. \\

For positive reward and optimal action, target will be greater than the current action value resulting in positive error which would result in $Q_{t+1}^\pi(s_t, a_t) > Q_{t}^\pi(s_t, a_t)$. As policy distribution is a function of action value, $\pi_{t+1}(s,a) > \pi_{t}(s,a)$, giving us $entropy(\pi_{t+1}(s)) < entropy(\pi_{t}(s))$. Hence, $confidence_{t+1}(s) > confidence_{t}(s)$. If we continue to train the RL algorithm, the confidence should be monotonically increasing.
For the limiting case,
\vspace{-3pt}
\begin{equation*}
    \lim\limits_{t \to \infty} \pi_{t}(s) \to 
    \begin{cases}
        \delta(s), & \text{continuous}  \\
        \delta_{a = \max_{a} Q_{t}^\pi(s_{t}, a)}, & \text{discrete}
    \end{cases}
\end{equation*}

where, $\delta(s)$ is the delta distribution of actions for a given state which is Dirac's delta distributionand $\delta$ is the Kronecker's delta. As entropy for a delta distribution is zero and given the definition of confidence in eq.\ref{eq::confidence}
\vspace{-2pt}
\begin{equation*}
    \lim\limits_{t \to \infty} \text{confidence}_{t}(s) \to 1
\end{equation*}
\end{proof}
This is also applicable when using functions approximation, where the target and error are captured in the loss,

\begin{equation*}
\label{eq::td_loss}
    \mathcal{L}(\theta_t) =  \mathbb{E}_{s_t,a_t,r_t,s_{t+1} \sim \mathcal{D}}[(\mathcal{R}_t + \gamma \max_{a} Q_{t}^\pi(s_{t+1}, a ; \theta_t) - Q_{t}^\pi(s_t, a_t ; \theta_t))^2]
\end{equation*}

Similarly, same can be shown for value iteration with Monte Carlo and TD($\lambda$) and policy iteration.

In RLInspect, action confidence is implemented for discrete action space, $\mathcal{A}$ and q-value is converted to policy distribution as
\vspace{-2pt}
\begin{equation}
\label{eq::q_to_policy_dist}
\begin{aligned}
    \pi(s,a_i) & = softmax(\text{q-value}) \\
             & = \frac{e^{q(s,a_i)}}{\sum_{a} e^{q(s,a)}} && \forall i = 1,2,\dots|\mathcal{A}|
\end{aligned}   
\end{equation}

And for calculating entropy, logarithm with base of size of action space is used.
\vspace{-2pt}
\begin{equation}
\label{eq::entropy}
    \text{entropy}(\pi(s)) = - \sum_{a} p(a|s)\log_{|\mathcal{A}|}p(a|s)
\end{equation}

where, $p(a|s)$ is the probability of action $\textit{a}$ given state $\textit{s}$.

\paragraph{Action convergence:}
Action for a given state converges to a particular action as the RL algorithm training converges. This analysis looks at how decision making of the RL algorithm evolves through training process.

\begin{proof}
    In eq.\ref{eq::q_update} as error $\to$ 0, $Q_{t+1}^\pi(s_t, a_t) \to Q_{t}^\pi(s_t, a_t)$ which follows $\pi_{t+1}(s, a) \to \pi_{t}(s, a)$. Same is applicable when using functions approximation, where $\theta_{t+1} \to \theta_{t}$ would correspond to $Q_{t+1}^\pi(s, a; \theta_{t+1}) \to Q_{t}^\pi(s, a; \theta_t)$.
\end{proof}

RLInspect is capable of evaluating this strategy for both continuous and discrete action space. This convergence is evaluated by calculating distance between action vectors from consecutive training updates
\vspace{-3pt}
\begin{equation}
    \text{change}_{t+1}^t = d(a_t, a_{t+1})
\end{equation}

where, $\text{change}_{t+1}^t$ is the change in training update \textit{t} and \textit{t+1} and $d(a_t, a_{t+1})$ is the distance between action vector at \textit{t} and \textit{t+1}. Euclidean distance is used for continuous action space and Jaccard distance in case of discrete action space.

\paragraph{Policy Divergence:}
This analysis focuses on changes in policy distribution between two training updates. To see how the policy is changing between two consecutive training updates, Jensen-Shannon divergence (JSD) \cite{jsd} with natural logarithm is calculated. JSD is chosen because firstly, it is well behaved for very small probabilities. Secondly, JSD with natural log has an upper bound of natural logarithm of 2 (ln(2)).

As the RL algorithm training converges, the divergence between the policy distribution of two consecutive training updates converges to zero.

\begin{proof}
    The proof for this continues from the proof for action convergence. As $Q_{t+1}^\pi(s_t, a_t) \to Q_{t}^\pi(s_t, a_t)$, $D(Q_{t+1}^\pi(s, a) || Q_{t}^\pi(s, a)) \to 0$. The policy distribution is calculated as defined in eq.\ref{eq::q_to_policy_dist}, which makes $D(\pi_{t+1}(s, a) || \pi_{t}(s, a)) \to 0$.
\end{proof}

This strategy can also identify the behaviour of action value update from eq.\ref{eq::q_update}. The divergence is governed by the change between action values from two consecutive training updates. From eq.\ref{eq::q_update}, it can be noted that the change in action values is attributed to learning rate, $\alpha$ and error.
\vspace{-2pt}
\begin{equation}
\label{eq::divengence}
    Q_{t+1}^\pi(s_t, a_t) - Q_{t}^\pi(s_t, a_t) = \alpha(\mathcal{R}_t + \gamma \max_{a} Q_{t}^\pi(s_{t+1}, a) - Q_{t}^\pi(s_t, a_t))
\end{equation}

If the divergence value is constantly high, then either of the two could contribute to this higher value. However, if the divergence sees a sudden spike and the learning rate, $\alpha$, is constant throughout the training, then it is due to higher error.

In RLInspect, this policy is implemented for discrete action space and entropy for divergence is calculated as defined in eq.\ref{eq::entropy}.

\begin{figure}[htb]
\begin{center}
    \includegraphics[width=0.9\textwidth, height=0.5\textwidth]{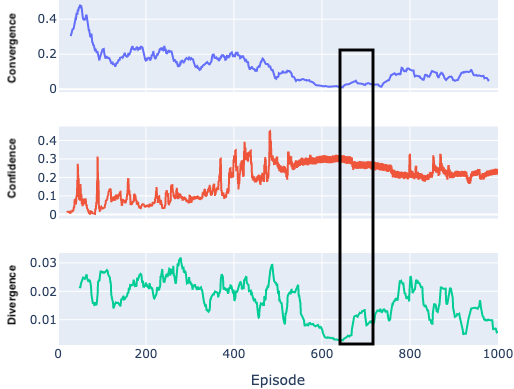}
    \caption{Convergence, Divergence and Confidence of Actions in Cartpole.}
    \label{fig: Convergence_Divergence_Confidence}
\end{center}
    
\end{figure}

On carrying out these analyses on the cartpole example the RL algorithm behaviour with respect to the action module can be seen in Figure.\ref{fig: Convergence_Divergence_Confidence}. It can be seen that episode 650 onwards there is drop in action confidence which also conincides with increase in action convergence. This behaviour could be due to an increase in policy divergence from the same episode. This increase indicates that there is significant change between $Q_{t+1}$ and $Q_{t}$ which from eq.\ref{eq::divengence} can be attributed to the error.

\vspace{-2pt}

\subsection{Agent Architecture Module}
\label{subsec:agent}
\begin{figure}[htb]
    \includegraphics[width=1\textwidth,height=0.6\textwidth]{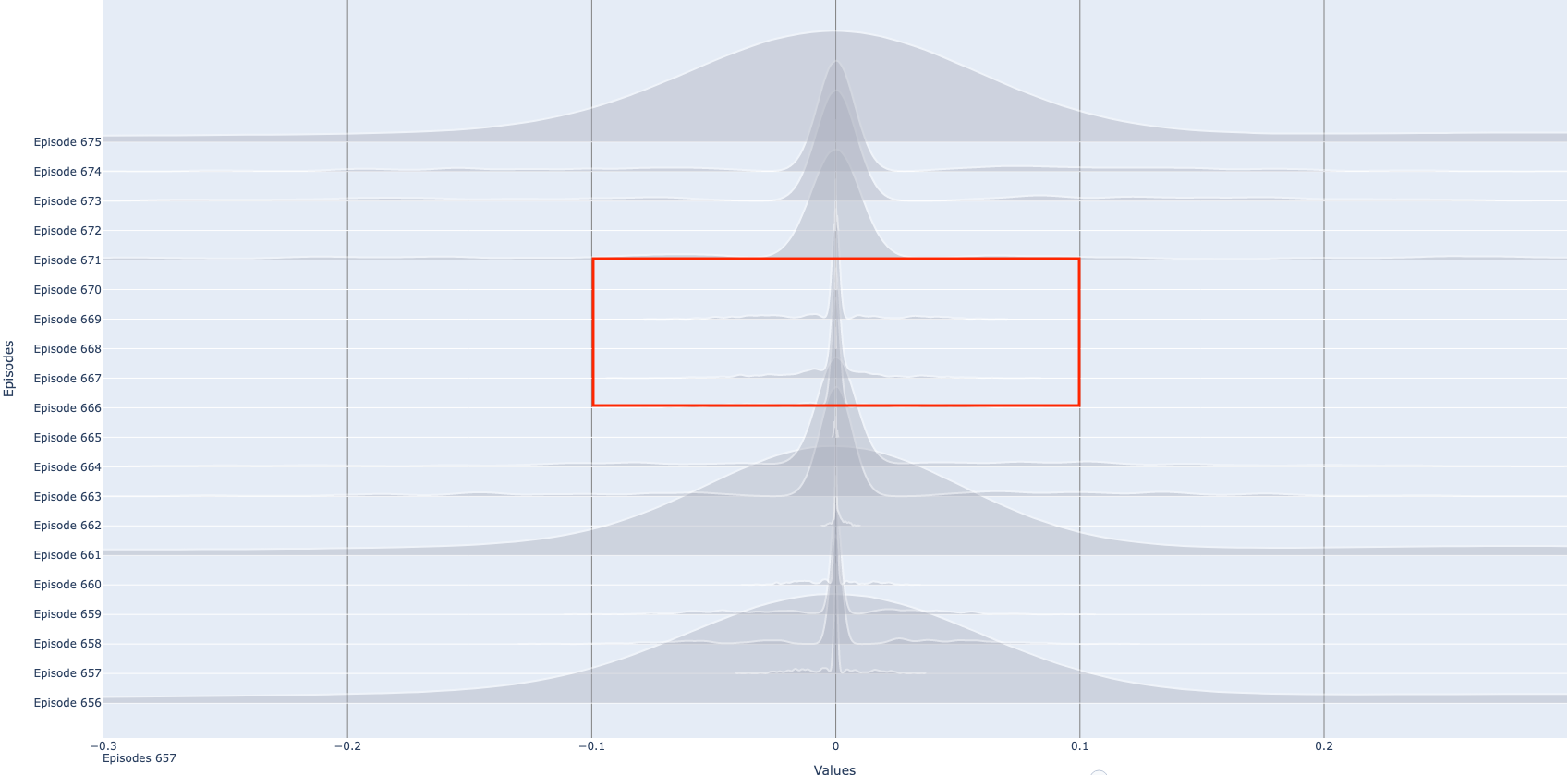}
    \caption{Gradient distribution for Cartpole. The red box in the figure shows the area where gradient became almost zero.}
    \label{fig: Gradient Distribution for Cartpole Environment}
\end{figure}

To monitor gradient-related issues that can hinder deep learning model performance, observing weight, bias, and gradients is crucial \cite{Glorot2010UnderstandingTD}. In deep reinforcement learning, where the agent's learning is guided by gradients, RLInspect's agent architecture module provides weight, bias, and gradient distribution analysis to monitor model performance. For instance, the gradient distribution analysis, as illustrated in Figure \ref{fig: Gradient Distribution for Cartpole Environment}, reveals the presence of the vanishing gradient problem for the cartpole example. This issue was observed during Episode 666-672, which coincided with a performance drop in the action module. By monitoring these parameters, users can make informed decisions to fine-tune their model and improve its overall performance.

\subsection{Reward Module}
\label{subsec:reward}
The reward module in RLInspect provides a comprehensive understanding of the behaviour of the reward function. As a first step, the module removes outliers from the data, which makes its analysis, such as average and exponential moving average per episode, more robust. In addition to these two analysis reward module offers two more analysis which include:

\paragraph{Volatility:}
Measuring the volatility in rewards can help in understanding the stability and consistency of the reward signal in reinforcement learning. High volatility in rewards means that the reward signal is inconsistent and may lead to unstable behaviour in the agent. On the other hand, low volatility indicates a stable and consistent reward signal, which can lead to better and more reliable behaviour from the agent.
\vspace{-2pt}
\begin{equation}
\sigma = \sqrt{\mathbb{E}[(r - \mathbb{E}[r])^2]}
\end{equation}

where $\sigma$ is the volatility and $r$ is the reward.

\begin{figure}[H]
    \includegraphics[width=0.9\textwidth]{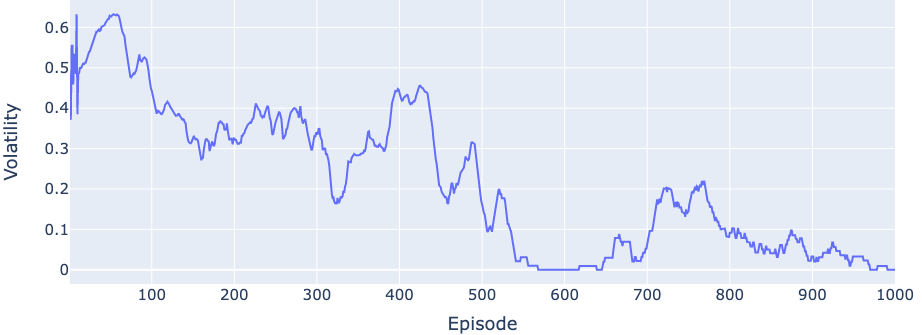}
    \caption{Volatility in rewards across episode for Cartpole.}
    \label{fig: Volatility in Rewards across episode for Cartpole Environment}
\end{figure}

As seen in the agent architecture module of the cartpole example, the vanishing gradient issue caused a significant drop in the model's performance around Episode 650-670. This drop in performance is also evident in the volatility of the rewards, as demonstrated in Figure \ref{fig: Volatility in Rewards across episode for Cartpole Environment}. The graph depicts an increase in reward volatility during this period, suggesting that the model is facing more uncertainty than the previous few episodes.

\paragraph{Risk-Reward Ratio:}
This analysis is for evaluating an agent's risk aversion. This represents the balance between the potential rewards and the risks it faces while taking actions. An agent with a higher risk-reward ratio would takes high risk to achieve the same reward compared to an agent with lower risk-reward ratio.

\begin{equation}
CV = \frac{\sigma}{ \mu } 
\end{equation}

where $\sigma$ is the standard deviation of the rewards per episode and $\mu$ is the mean of the rewards per episode.

% \subsection{Loss Module}
% \label{subsec:loss}
% By computing the difference between predicted and actual values, the loss function provides a signal that can be used to update the parameters of the policy in a way that minimizes the difference between the two. Monitoring the loss function during the training process is crucial because it provides crucial feedback to the agent regarding its learning progress. To facilitate this monitoring process, Loss Module generates a plot of the loss function over time. This plot enables users to visualize how well the agent is learning and identify any potential issues or inefficiencies in the learning process.

\subsection{Extendibility}
\label{subsec:extendibility}

RLInspect is developed considering its extendibility beyond the modules, analyses and data handler that are available by default.
% This can be achived through following ways:
% \begin{itemize}
%     \item The user can carry out their custom analysis in a specific module by creating their own analysis by implementing \texttt{analyse} and \texttt{plot} methods.
%     \item The user can extend the base \texttt{Analyzer} to create their own custom analyser and use it to collect data of their choosing.
%     \item The user can create their custom data handler which may suite their requirements and can potentially act as data transformer during IO opertations.
% \end{itemize}
The user can perform custom analysis by extending or implementing their own \texttt{analyse} and \texttt{plot} methods.
If the user wants to collect data of their choosing and carry out analysis, the base \texttt{Analyzer} can be extended to create a custom analyser. Further, the user can create their custom data handler which may suit their requirements and can potentially act as the data transformation layer during IO operations.

 % RLInspect is developed in such a way that  The data collection is a crutial part of the analysis and the volume and diversity of data may grow depending on the type of data being collected and analyses being run. Therefore, to accommodate diversity and type,  This custom data handler can then be provided to RLInspect to read and write data and can potentially act as data transformer during IO opertations.
% \input{content/case-study}
\section{Limitations and Future works}
\label{sec::future_works}

\textbf{Dimensionality reduction:} The state module uses dimensionality reduction to embed high dimensional state onto two dimensional space for visualisation. RLInspect uses incremental PCA for embedding high dimentional state vectors to two dimensions. However, it suffers from its own drawbacks \cite{aupetit2007visualizing,heulot2013proxilens,aupetit2014visualisation}. Another candidate for dimensionality reduction was metric multidimensional scaling (mMDS) \cite{mds_review} but it was soon discarded as it becomes computationally expensive for large data. Therefore, other dimensionality reduction algorithms like t-sne \cite{tsne2008vandermaaten08a} needs to be explored as alternative solutions.

\textbf{Action confidence for continuous action space:} Currently action module is able to calculate action confidence and policy divergence only for discrete action space. As future work, these will be extended for continuous action space.
\section{Conclusion}
\label{sec::conslusion}

In this paper, we introduced RLInspect - an intteractive visual analytics tool, that provides comprehensive understanding of RL training and can potentially help in improving RL algorithm. The tool looks at four major components of RL training i.e. state, action, reward and agent architecture, to assess the overall behaviour of an RL algorithm. The modular design of RLInspect provides user an option to choose what analysis to run. The extendability aspect of RLInspect enables user to define their custom analysis and data handler which can further help in a much more detailed analysis of the RL algorithm.

\bibliographystyle{unsrtnat}
\bibliography{document}

\begin{thebibliography}{40}
\providecommand{\natexlab}[1]{#1}
\providecommand{\url}[1]{\texttt{#1}}
\expandafter\ifx\csname urlstyle\endcsname\relax
  \providecommand{\doi}[1]{doi: #1}\else
  \providecommand{\doi}{doi: \begingroup \urlstyle{rm}\Url}\fi

\bibitem[LeCun et~al.(2015)LeCun, Bengio, and Hinton]{LeCun2015-yj}
Yann LeCun, Yoshua Bengio, and Geoffrey Hinton.
\newblock Deep learning.
\newblock \emph{Nature}, 521\penalty0 (7553):\penalty0 436--444, May 2015.

\bibitem[Blagec et~al.(2021)Blagec, Dorffner, Moradi, and
  Samwald]{blagec2021critical}
Kathrin Blagec, Georg Dorffner, Milad Moradi, and Matthias Samwald.
\newblock A critical analysis of metrics used for measuring progress in
  artificial intelligence, 2021.

\bibitem[Zhou et~al.(2021)Zhou, Gandomi, Chen, and Holzinger]{Zhou2021-fd}
Jianlong Zhou, Amir~H Gandomi, Fang Chen, and Andreas Holzinger.
\newblock Evaluating the quality of machine learning explanations: A survey on
  methods and metrics.
\newblock \emph{Electronics}, 10\penalty0 (5), 2021.

\bibitem[Wongsuphasawat et~al.(2018)Wongsuphasawat, Smilkov, Wexler, Wilson,
  Mane, Fritz, Krishnan, Viegas, and Wattenberg]{dataflow_graph_tensorflow}
Kanit Wongsuphasawat, Daniel Smilkov, James Wexler, Jimbo Wilson, Dandelion
  Mane, Doug Fritz, Dilip Krishnan, Fernanda~B. Viegas, and Martin Wattenberg.
\newblock Visualizing dataflow graphs of deep learning models in tensorflow.
\newblock \emph{IEEE Transactions on Visualization and Computer Graphics},
  24\penalty0 (1):\penalty0 1–12, 2018.
\newblock \doi{10.1109/tvcg.2017.2744878}.

\bibitem[Satyanarayan et~al.(2017)Satyanarayan, Moritz, Wongsuphasawat, and
  Heer]{vega_lite}
Arvind Satyanarayan, Dominik Moritz, Kanit Wongsuphasawat, and Jeffrey Heer.
\newblock Vega-lite: A grammar of interactive graphics.
\newblock \emph{IEEE Transactions on Visualization and Computer Graphics},
  23\penalty0 (1):\penalty0 341--350, 2017.
\newblock \doi{10.1109/TVCG.2016.2599030}.

\bibitem[Kahng et~al.(2017)Kahng, Andrews, Kalro, and Chau]{kahng2017cti}
Minsuk Kahng, Pierre~Y Andrews, Aditya Kalro, and Duen~Horng Chau.
\newblock Activis: Visual exploration of industry-scale deep neural network
  models.
\newblock \emph{IEEE transactions on visualization and computer graphics},
  24\penalty0 (1):\penalty0 88--97, 2017.

\bibitem[Krause et~al.(2016)Krause, Perer, and
  Ng]{interacting_with_predictions}
Josua Krause, Adam Perer, and Kenney Ng.
\newblock Interacting with predictions: Visual inspection of black-box machine
  learning models.
\newblock In \emph{Proceedings of the 2016 CHI Conference on Human Factors in
  Computing Systems}, CHI '16, page 5686–5697, New York, NY, USA, 2016.
  Association for Computing Machinery.
\newblock ISBN 9781450333627.
\newblock \doi{10.1145/2858036.2858529}.
\newblock URL \url{https://doi.org/10.1145/2858036.2858529}.

\bibitem[Yosinski et~al.(2015)Yosinski, Clune, Nguyen, Fuchs, and
  Lipson]{yosinski2015understanding}
Jason Yosinski, Jeff Clune, Anh Nguyen, Thomas Fuchs, and Hod Lipson.
\newblock Understanding neural networks through deep visualization.
\newblock \emph{arXiv preprint arXiv:1506.06579}, 2015.

\bibitem[Pugliese et~al.(2021)Pugliese, Regondi, and Marini]{rl_popularity}
Raffaele Pugliese, Stefano Regondi, and Riccardo Marini.
\newblock Machine learning-based approach: global trends, research directions,
  and regulatory standpoints.
\newblock \emph{Data Science and Management}, 4:\penalty0 19--29, 2021.
\newblock ISSN 2666-7649.
\newblock \doi{https://doi.org/10.1016/j.dsm.2021.12.002}.
\newblock URL
  \url{https://www.sciencedirect.com/science/article/pii/S2666764921000485}.

\bibitem[Li(2019)]{li2019reinforcement}
Yuxi Li.
\newblock Reinforcement learning applications.
\newblock \emph{arXiv preprint arXiv:1908.06973}, 2019.

\bibitem[Lambert et~al.(2022)Lambert, Castricato, von Werra, and
  Havrilla]{chatgpt_rlhf}
Nathan Lambert, Louis Castricato, Leandro von Werra, and Alex Havrilla.
\newblock Illustrating reinforcement learning from human feedback (rlhf).
\newblock \emph{Hugging Face Blog}, 2022.
\newblock https://huggingface.co/blog/rlhf.

\bibitem[Liu et~al.(2023)Liu, Han, Ma, Zhang, Yang, Tian, He, Li, He, Liu,
  et~al.]{liu2023summary}
Yiheng Liu, Tianle Han, Siyuan Ma, Jiayue Zhang, Yuanyuan Yang, Jiaming Tian,
  Hao He, Antong Li, Mengshen He, Zhengliang Liu, et~al.
\newblock Summary of chatgpt/gpt-4 research and perspective towards the future
  of large language models.
\newblock \emph{arXiv preprint arXiv:2304.01852}, 2023.

\bibitem[Zheng et~al.(2022)Zheng, Trott, Srinivasa, Parkes, and
  Socher]{ai_economist}
Stephan Zheng, Alexander Trott, Sunil Srinivasa, David~C. Parkes, and Richard
  Socher.
\newblock The ai economist: Taxation policy design via two-level deep
  multiagent reinforcement learning.
\newblock \emph{Science Advances}, 8\penalty0 (18), 2022.
\newblock \doi{10.1126/sciadv.abk2607}.

\bibitem[Tilbury(2022)]{tilbury2022reinforcement}
Callum Tilbury.
\newblock Reinforcement learning in macroeconomic policy design: A new
  frontier?
\newblock \emph{arXiv preprint arXiv:2206.08781}, 2022.

\bibitem[Kwak et~al.(2021)Kwak, Ling, and Hui]{kwak_ling_hui_2021}
Gloria~Hyunjung Kwak, Lowell Ling, and Pan Hui.
\newblock Deep reinforcement learning approaches for global public health
  strategies for covid-19 pandemic.
\newblock \emph{PLOS ONE}, 16\penalty0 (5), 2021.
\newblock \doi{10.1371/journal.pone.0251550}.

\bibitem[Khadilkar et~al.(2020)Khadilkar, Ganu, and
  Seetharam]{khadilkar2020optimising}
Harshad Khadilkar, Tanuja Ganu, and Deva~P Seetharam.
\newblock Optimising lockdown policies for epidemic control using reinforcement
  learning: An ai-driven control approach compatible with existing disease and
  network models.
\newblock \emph{Transactions of the Indian National Academy of Engineering},
  5\penalty0 (2):\penalty0 129--132, 2020.

\bibitem[Hambly et~al.(2021)Hambly, Xu, and Yang]{hambly2021recent}
Ben Hambly, Renyuan Xu, and Huining Yang.
\newblock Recent advances in reinforcement learning in finance.
\newblock \emph{arXiv preprint arXiv:2112.04553}, 2021.

\bibitem[Liu et~al.(2020)Liu, See, Ngiam, Celi, Sun, and
  Feng]{rl_clinical_decision_system}
Siqi Liu, Kay~Choong See, Kee~Yuan Ngiam, Leo~Anthony Celi, Xingzhi Sun, and
  Mengling Feng.
\newblock Reinforcement learning for clinical decision support in critical
  care: Comprehensive review.
\newblock \emph{Journal of Medical Internet Research}, 22\penalty0 (7), 2020.
\newblock \doi{10.2196/18477}.

\bibitem[Van~Seijen et~al.(2020)Van~Seijen, Nekoei, Racah, and
  Chandar]{NEURIPS2020_48db7158}
Harm Van~Seijen, Hadi Nekoei, Evan Racah, and Sarath Chandar.
\newblock The loca regret: A consistent metric to evaluate model-based behavior
  in reinforcement learning.
\newblock In H.~Larochelle, M.~Ranzato, R.~Hadsell, M.F. Balcan, and H.~Lin,
  editors, \emph{Advances in Neural Information Processing Systems}, volume~33,
  pages 6562--6572. Curran Associates, Inc., 2020.
\newblock URL
  \url{https://proceedings.neurips.cc/paper_files/paper/2020/file/48db71587df6c7c442e5b76cc723169a-Paper.pdf}.

\bibitem[Jordan et~al.(2020)Jordan, Chandak, Cohen, Zhang, and
  Thomas]{jordan2020evaluating}
Scott Jordan, Yash Chandak, Daniel Cohen, Mengxue Zhang, and Philip Thomas.
\newblock Evaluating the performance of reinforcement learning algorithms.
\newblock In \emph{International Conference on Machine Learning}, pages
  4962--4973. PMLR, 2020.

\bibitem[Tufte(2018)]{tufte_2018}
Edward~R. Tufte.
\newblock \emph{The visual display of quantitative information}.
\newblock Graphics Press, 2018.

\bibitem[Yi et~al.(2007)Yi, Kang, Stasko, and Jacko]{jisoo2007interaction}
Ji~Soo Yi, Youn~ah Kang, John Stasko, and J.A. Jacko.
\newblock Toward a deeper understanding of the role of interaction in
  information visualization.
\newblock \emph{IEEE Transactions on Visualization and Computer Graphics},
  13\penalty0 (6):\penalty0 1224--1231, 2007.
\newblock \doi{10.1109/TVCG.2007.70515}.

\bibitem[van Wesel and Goodloe(2017)]{Wesel2017ChallengesIT}
Perry van Wesel and Alwyn Goodloe.
\newblock Challenges in the verification of reinforcement learning algorithms.
\newblock 2017.

\bibitem[Iyer et~al.(2018)Iyer, Li, Li, Lewis, Sundar, and
  Sycara]{iyer2018obj_saliency_map}
Rahul Iyer, Yuezhang Li, Huao Li, Michael Lewis, Ramitha Sundar, and Katia
  Sycara.
\newblock Transparency and explanation in deep reinforcement learning neural
  networks.
\newblock In \emph{Proceedings of the 2018 AAAI/ACM Conference on AI, Ethics,
  and Society}, AIES '18, page 144–150, New York, NY, USA, 2018. Association
  for Computing Machinery.
\newblock ISBN 9781450360128.
\newblock \doi{10.1145/3278721.3278776}.
\newblock URL \url{https://doi.org/10.1145/3278721.3278776}.

\bibitem[Chan et~al.(2020)Chan, Fishman, Canny, Korattikara, and
  Guadarrama]{chan2020measuring}
Stephanie C.~Y. Chan, Samuel Fishman, John Canny, Anoop Korattikara, and Sergio
  Guadarrama.
\newblock Measuring the reliability of reinforcement learning algorithms, 2020.

\bibitem[Sequeira and Gervasio(2020)]{Sequeira_2020}
Pedro Sequeira and Melinda Gervasio.
\newblock Interestingness elements for explainable reinforcement learning:
  Understanding agents{\textquotesingle} capabilities and limitations.
\newblock \emph{Artificial Intelligence}, 288:\penalty0 103367, nov 2020.
\newblock \doi{10.1016/j.artint.2020.103367}.
\newblock URL \url{https://doi.org/10.1016\%2Fj.artint.2020.103367}.

\bibitem[Feit et~al.(2022)Feit, Metzger, and Pohl]{feit2022explaining}
Felix Feit, Andreas Metzger, and Klaus Pohl.
\newblock Explaining online reinforcement learning decisions of self-adaptive
  systems.
\newblock In \emph{2022 IEEE International Conference on Autonomic Computing
  and Self-Organizing Systems (ACSOS)}, pages 51--60. IEEE, 2022.

\bibitem[McGregor et~al.(2017)McGregor, Buckingham, Dietterich, Houtman,
  Montgomery, and Metoyer]{MCGREGOR201793}
Sean McGregor, Hailey Buckingham, Thomas~G. Dietterich, Rachel Houtman, Claire
  Montgomery, and Ronald Metoyer.
\newblock Interactive visualization for testing markov decision processes:
  Mdpvis.
\newblock \emph{Journal of Visual Languages \& Computing}, 39:\penalty0
  93--106, 2017.
\newblock ISSN 1045-926X.
\newblock \doi{https://doi.org/10.1016/j.jvlc.2016.10.007}.
\newblock URL
  \url{https://www.sciencedirect.com/science/article/pii/S1045926X16301951}.
\newblock Special Issue on Programming and Modelling Tools.

\bibitem[Wang et~al.(2019)Wang, Gou, Shen, and Yang]{wang2019dqnviz}
Junpeng Wang, Liang Gou, Han-Wei Shen, and Hao Yang.
\newblock Dqnviz: A visual analytics approach to understand deep q-networks.
\newblock \emph{IEEE Transactions on Visualization and Computer Graphics},
  25\penalty0 (1):\penalty0 288--298, 2019.
\newblock \doi{10.1109/TVCG.2018.2864504}.

\bibitem[Jaunet et~al.(2020)Jaunet, Vuillemot, and Wolf]{jaunet2020drlviz}
Theo Jaunet, Romain Vuillemot, and Christian Wolf.
\newblock Drlviz: Understanding decisions and memory in deep reinforcement
  learning.
\newblock In \emph{Computer Graphics Forum}, volume~39, pages 49--61. Wiley
  Online Library, 2020.

\bibitem[Inc.(2015)]{plotly}
Plotly~Technologies Inc.
\newblock Collaborative data science, 2015.
\newblock URL \url{https://plot.ly}.

\bibitem[Brockman et~al.(2016)Brockman, Cheung, Pettersson, Schneider,
  Schulman, Tang, and Zaremba]{brockman2016openai}
Greg Brockman, Vicki Cheung, Ludwig Pettersson, Jonas Schneider, John Schulman,
  Jie Tang, and Wojciech Zaremba.
\newblock Openai gym, 2016.

\bibitem[Ross et~al.(2007)Ross, Lim, Lin, and Yang]{Ross2007}
David~A. Ross, Jongwoo Lim, Ruei-Sung Lin, and Ming-Hsuan Yang.
\newblock Incremental learning for robust visual tracking.
\newblock \emph{International Journal of Computer Vision}, 77\penalty0
  (1-3):\penalty0 125--141, August 2007.
\newblock \doi{10.1007/s11263-007-0075-7}.
\newblock URL \url{https://doi.org/10.1007/s11263-007-0075-7}.

\bibitem[Lin(1991)]{jsd}
J.~Lin.
\newblock Divergence measures based on the shannon entropy.
\newblock \emph{IEEE Transactions on Information Theory}, 37\penalty0
  (1):\penalty0 145--151, 1991.
\newblock \doi{10.1109/18.61115}.

\bibitem[Glorot and Bengio(2010)]{Glorot2010UnderstandingTD}
Xavier Glorot and Yoshua Bengio.
\newblock Understanding the difficulty of training deep feedforward neural
  networks.
\newblock In \emph{International Conference on Artificial Intelligence and
  Statistics}, 2010.

\bibitem[Aupetit(2007)]{aupetit2007visualizing}
Micha{\"e}l Aupetit.
\newblock Visualizing distortions and recovering topology in continuous
  projection techniques.
\newblock \emph{Neurocomputing}, 70\penalty0 (7-9):\penalty0 1304--1330, 2007.

\bibitem[Heulot et~al.(2013)Heulot, Aupetit, and Fekete]{heulot2013proxilens}
N.~Heulot, M.~Aupetit, and J-D. Fekete.
\newblock {ProxiLens: Interactive Exploration of High-Dimensional Data using
  Projections}.
\newblock In M.~Aupetit and L.~van~der Maaten, editors, \emph{EuroVis Workshop
  on Visual Analytics using Multidimensional Projections}. The Eurographics
  Association, 2013.
\newblock ISBN 978-3-905674-53-8.
\newblock \doi{10.2312/PE.VAMP.VAMP2013.011-015}.

\bibitem[Aupetit et~al.(2014)Aupetit, Heulot, and
  Fekete]{aupetit2014visualisation}
Michael Aupetit, Nicolas Heulot, and Jean-Daniel Fekete.
\newblock A multidimensional brush for scatterplot data analytics.
\newblock In \emph{2014 IEEE Conference on Visual Analytics Science and
  Technology (VAST)}, pages 221--222, 2014.
\newblock \doi{10.1109/VAST.2014.7042500}.

\bibitem[Mead(2018)]{mds_review}
A.~Mead.
\newblock {Review of the Development of Multidimensional Scaling Methods}.
\newblock \emph{Journal of the Royal Statistical Society Series D: The
  Statistician}, 41\penalty0 (1):\penalty0 27--39, 12 2018.
\newblock ISSN 2515-7884.
\newblock \doi{10.2307/2348634}.
\newblock URL \url{https://doi.org/10.2307/2348634}.

\bibitem[van~der Maaten and Hinton(2008)]{tsne2008vandermaaten08a}
Laurens van~der Maaten and Geoffrey Hinton.
\newblock Visualizing data using t-sne.
\newblock \emph{Journal of Machine Learning Research}, 9\penalty0
  (86):\penalty0 2579--2605, 2008.
\newblock URL \url{http://jmlr.org/papers/v9/vandermaaten08a.html}.

\end{thebibliography}
%%%%%%%%%%%%%%%%%%%%%%%%%%%%%%%%%%%%%%%%%%%%%%%%%%%%%%%%%%%%

\end{document}